\documentclass{article}

\usepackage{PRIMEarxiv}

\usepackage[utf8]{inputenc} 
\usepackage[T1]{fontenc}    
\usepackage{hyperref}       
\usepackage{url}            
\usepackage{booktabs}       
\usepackage{amsfonts}       
\usepackage{nicefrac}       
\usepackage{microtype}      
\usepackage{lipsum}
\usepackage{fancyhdr}       
\usepackage{graphicx}       
\graphicspath{{media/}}     

\usepackage{multirow}  
\usepackage{amsmath}   
\usepackage{adjustbox}  
\usepackage{amssymb}

\title{Zero-Shot CFC: Fast Real-World Image Denoising based on Cross-Frequency Consistency} 

\author{
  Yanlin Jiang \\
  School of Information Science and Technology \\
  Beijing University of Technology \\
  \texttt{SegelJiang@gmail.com} \\
   \And
  Yuchen Liu \\
  School of Information Science and Technology \\
  Beijing University of Technology \\
  \texttt{lyc0118@emails.bjut.edu.cn} \\
  \AND
  Mingren Liu \\
  Alibaba Cloud \\
  \texttt{liumingren.lmr@alibaba-inc.com} \\
}

\begin{document}
\maketitle

\begin{abstract}
Zero-shot denoisers address the dataset dependency of deep-learning-based denoisers, enabling the denoising of unseen single images. Nonetheless, existing zero-shot methods suffer from long training times and rely on the assumption of noise independence and a zero-mean property, limiting their effectiveness in real-world denoising scenarios where noise characteristics are more complicated. This paper proposes an efficient and effective method for real-world denoising, the Zero-Shot denoiser based on Cross-Frequency Consistency (ZSCFC), which enables training and denoising with a single noisy image and does not rely on assumptions about noise distribution. Specifically, image textures exhibit position similarity and content consistency across different frequency bands, while noise does not. Based on this property, we developed cross-frequency consistency loss and an ultralight network to realize image denoising. Experiments on various real-world image datasets demonstrate that our ZSCFC outperforms other state-of-the-art zero-shot methods in terms of computational efficiency and denoising performance.
\end{abstract}


\section{Introduction}
Image noise can degrade overall image quality, resulting in reduced clarity, color distortion, loss of textures, and introduction of compression artifacts \cite{p1_nature_bio, p1_Liu_2023_ICCV, p1_Zou_2023_ICCV}. Image denoising aims to eliminate noise while preserving critical textures and underlying structures, presenting a challenge in balancing effective noise reduction with the preservation of fine textures and essential features. In real-world conditions, noise intensity and distribution vary randomly \cite{rwn_Cheng_2023_ICCV, rwn_Lin_2023_ICCV, rwn_Wang_2023_ICCV}. For example, noise may randomly concentrate in certain regions, making it harder for denoisers to adapt uniformly across the image without sacrificing important textures.

Current supervised image denoising methods \cite{s1_dncnn, s2_ffdnet, s3_restormer, s4} and self-supervised denoising methods \cite{self1_n2v, self2_ne2ne, self3_lgbpn, self4_sdap, self5_b2unb} require large amounts of noisy training data to achieve high denoising performance. To address limitations of dataset requirements, several zero-shot/dataset-free methods \cite{z1_DIP, z2_S2S, z3_N2F, z4_ZSN2N, z5_nc} have recently been developed to perform real-world denoising using only a single noisy image. Most of them are grounded in the Noise2Noise theoretical framework, which suggests that when independently distributed noise has a zero mean, training a network to map one noisy image to another noisy image of the same scene can yield results comparable to using clean ground-truth images. However, these methods typically require splitting a single noisy image to create noisy/noisy subimage pairs, disrupting internal relationships between neighboring pixels in the spatial domain.

Given these challenges, it is essential to develop a zero-shot denoising method for real-world images that can maximally protect the original information of a noisy image and overcome the randomness of real-world noise distributions to achieve effective denoising. Inspired by prior studies \cite{fre1, fre2}, we observe high-frequency edges and textures are supported by the underlying structures and primary colors, exhibiting consistency across multiple frequency bands, and demonstrating structured and natural characteristics. In contrast, noise is randomly distributed across different high-frequency bands and lacks coherence. Additionally, according to \cite{z1_DIP}, the low impedance of the network further enhances its ability to learn structured and natural content, but it struggles to capture the irregular characteristics of real-world noise. On the basis of this observation, a network can be trained to extract fine textures from the high-frequency bands of an image. Inspired by these theories, we propose a fast Zero-Shot denoiser based on Cross-Frequency Consistency (ZSCFC).

The ZSCFC first decomposes a noisy image into multiple frequency bands. Then an ultralight network is designed as a texture extractor, learning image texture features based on the consistency of high-frequency information across multiple high-frequency bands. The high-frequency texture extractor needs to capture only the structured high-frequency textures from these subimages, without learning the underlying structure and colors of the image, thus achieving superior results with a minimal network size. Additionally, the ZSCFC shows strong robustness in handling complicated real-world noise. Experimental results on several real-world image datasets show that the ZSCFC outperforms other recent dataset-free methods in both computational efficiency and denoising performance.

Our main contributions are summarized as follows: \begin{itemize}
\item We design ZSCFC, a novel zero-shot method for real-world image denoising based on our proposed cross-frequency consistency loss without any noise model assumptions, guiding the network to realize the effective texture restoration which is the most challenging objective in denoising tasks.
\item We propose an ultralight network with only 1.5k parameters and 3s GPU denoising time for a single noisy image but can outperform the larger networks with millions of parameters, which is suitable for use on the edge device with limited computational resources.
\item Our method has outperformed state-of-the-art (SOTA) self-supervised and zero-shot denoising methods on real-world image datasets in computational efficiency and denoising performance, which shows its potential applications in real-world scenarios. 
\end{itemize}

\section{Related Work}
\label{sec:related_work}

\textbf{Supervised methods} Supervised denoising methods \cite{s1_dncnn, s2_ffdnet, s3_restormer, s4} achieve high-quality results by using paired noisy-clean images for end-to-end training, often employing complex architectures like CNNs and transformers to model noise patterns effectively. These methods  excel at capturing multiscale features and significantly outperform traditional methods like NLM \cite{tr3_nlm}, BM3D \cite{tr1_bm3d}, and WNNM \cite{tr4_wnnm}. However, their performance relies heavily on large, well-aligned noisy/clean datasets, which are costly and difficult to collect in real-world scenarios. Moreover, models trained on synthetic noise often fail to generalize to real-world scenarios due to domain gaps, limiting their practical use.

\noindent\textbf{Self-supervised methods} Self-supervised denoising methods \cite{self1_n2v, self2_ne2ne, self3_lgbpn, self4_sdap, self5_b2unb} rely solely on noisy images for network training. 
Assuming independent noisy pixels, Neighbor2Neighbor (Ne2Ne) \cite{self2_ne2ne} simplifies sample pair generation by creating training image pairs via a random neighbor sub-sampler. 
Noise2Void (N2V) \cite{self1_n2v} employs a blind-spot network (BSN) that masks the central pixel of each receptive field, using surrounding pixels for prediction to avoid identity mapping. 
Local and global blind-patch network (LG-BPN) \cite{self3_lgbpn} improved the masked scheme by leveraging the correlation statistic to realize a denser local receptive field and introduced a dilated Transformer block to allow exploitation of the distant context exploitation in the BSN. 
Sampling Difference As Perturbation (SDAP) \cite{self4_sdap} proposes a self-supervised denoising framework based on Random Sub-samples Generation to improve the performance of BSN by adding an appropriate perturbation to the training images. Despite these advancements, self-supervised denoising methods remain limited by reliance on specific noise models or assumptions and data acquisition challenges.

\noindent\textbf{Zero-shot methods} Zero-shot denoising methods \cite{z1_DIP, z2_S2S, z3_N2F, z4_ZSN2N, z5_nc} are designed to perform denoising without relying on clean images or large datasets, typically using only a single noisy image to train the network. 
Deep Image Prior (DIP) \cite{z1_DIP} uses a randomly initialized neural network to approximate a noisy image leveraging the inductive bias of the network to distinguish between noise and the underlying image structure. However, DIP is sensitive to the number of training iterations, requiring careful control to avoid overfitting. 
Self2Self (S2S) \cite{z2_S2S} deploys the Bernoulli-sampled strategy to create input training pairs and derives the denoising output by averaging the predictions generated from multiple instances of the trained model with dropout. 
Based on Noise2Noise theory, Noise2Fast (N2F) \cite{z3_N2F} utilizes a checkerboard downsampling to produce a four-image dataset for training, though it still requires spatially independent noise assumptions. 
Zero-shot noise2noise (ZSN2N) \cite{z4_ZSN2N} extends the zero-shot approach by applying fixed filters to a noisy test image, generating two corresponding downsampled versions to create input-target pairs and training a lightweight network on this pair without any training dataset. 

However, except for specific noise models or assumptions reliance limitation, the process of creating training pairs can disrupt the spatial consistency within the image, potentially compromising texture and structure and then resulting in suboptimal noise reduction quality. Furthermore, zero-shot methods require long training times, making them impractical for deployment on edge devices with limited computational resources in real-world scenarios.

\section{Method}
\label{sec:formatting}
\subsection{Overview}
\begin{figure*}[t]
  \centering
  \includegraphics[width=0.7\textwidth]{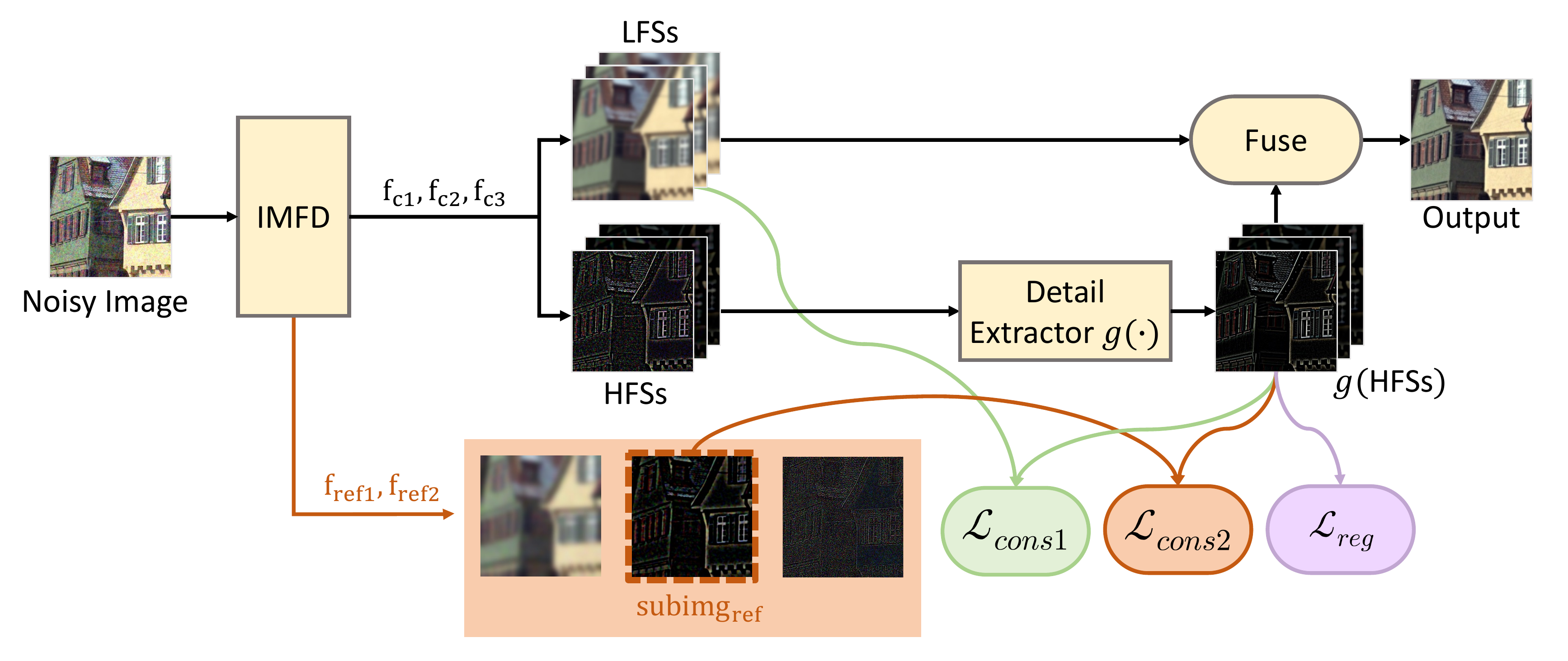}
  \caption{\textbf{The overall architecture of ZSCFC.} }
  \label{fig: mainArch}
\end{figure*}

The proposed method ZSCFC is a zero-shot method capable of denoising a single noisy image. 
This method first deploys the Image Multi-Frequency Decomposer (IMFD) to iteratively decompose the noisy image to one low-frequency subimage (LFS) and three high-frequency subimages (HFSs), denoted $\mathrm{LFS_1}$, $\mathrm{HFS_1}$, $\mathrm{HFS_2}$, and $\mathrm{HFS_3}$, with increasing frequency content (Sec \ref{sec: Decomposer}). Due to the low-frequency nature of $\mathrm{LFS_1}$, it contains almost no texture information or noise, therefore $\mathrm{LFS_1}$ is kept unchanged to maximize the retention of the underlying structure of the image. Then, an ultralight network with only 1.5k parameters $\mathrm{g(\cdot )}$ is employed as a texture extractor to fetch texture from $\mathrm{HFSs}$ (Sec \ref{sec: Network}). This network is guided by our proposed cross-frequency consistency loss (Sec \ref{sec: 3.3}). Finally, $\mathrm{LFS_1}$ that contains the image’s underlying structure is fused with the extracted texture from $\mathrm{HFSs}$ to generate the denoised image:


\begin{equation}
\mathrm{img_{denoi}\!=\!LFS_{1}\!+\!g(HFS_{1})\!+\!g(HFS_{2})\!+\!g(HFS_{3})}
\end{equation}
The illustration of the overall architecture of ZSCFC is in Fig. \ref{fig: mainArch}.

\subsection{Preliminary}
\label{sec: Preliminary}
An image can be separated into a $\mathrm{LFS}$ and a $\mathrm{HFS}$ through frequency decomposition. The LFS contains the underlying structures and basic colors of the image, with minimal noise. In contrast, the HFS includes image textures, such as edges and details, where also most of the noise in the noisy image is concentrated. The examples of LFSs and HFSs can be seen in Fig. \ref{fig: frequency decomposer}. Therefore, to enhance the overall consistency and fidelity of the denoised image, we perform denoising exclusively on HFS, thereby preserving the primary structure in LFS while maximizing the restoration of textures in HFS.

To obtain a pair of LFS and HFS from an image, we use ${\mathrm{f}_c}$ to set up a Gaussian kernel for conducting image frequency decomposition. ${\mathrm{f}_{c}}$ can be used to calculate the corresponding $\sigma$ through the formula $\sigma=\frac{1}{2\pi {\mathrm{f}_c}}$. Based on the calculated $\sigma$, a Gaussian kernel can be designed to derive the LFS \cite{fre1}. This kernel utilizes $\sigma$ and the nearest odd integer obtained by rounding up $6\sigma$ as the kernel size $\mathrm{k}$. The rationale behind using $6\sigma$ is that the range of $\pm 3\sigma $ from the mean in a Gaussian distribution contains $99.73\%$ of the information, thus minimizing the loss of data.

\begin{figure}[t]
  \centering
  \includegraphics[width=0.7\textwidth]{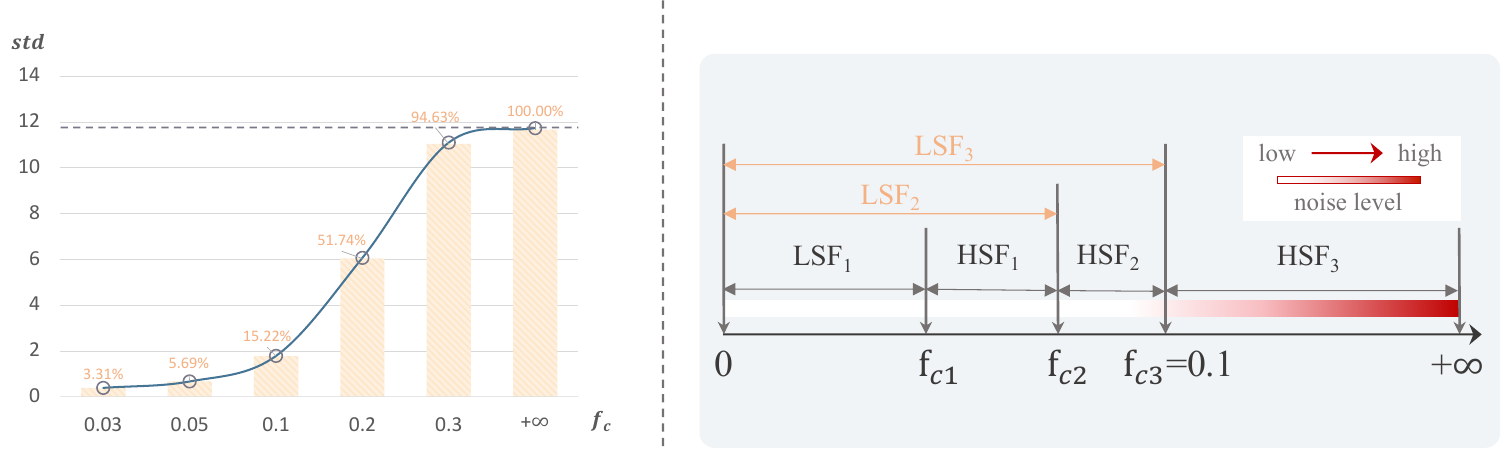}
  \caption{\textbf{(left)} Average LFS residual noise $\mathrm{std}$ at different $\mathrm{f_c}$. \textbf{(right)} The frequency bands of HFSs and LFSs.}
  \label{fig: 31low_freq_noise}
\end{figure}

To determine the optimal cutoff frequency ${\mathrm{f}_{c}}$, we conducted an analysis to measure the residual noise present in the LFS after frequency decomposition. We apply ${\mathrm{f}_{c}}$ to perform frequency decomposition on images from the SIDD Medium dataset and calculate the average $\mathrm{std}$ of residual noise in the $\mathrm{LFS}$, as shown in Fig. \ref{fig: 31low_freq_noise} (left). It can be seen that the average $\mathrm{std}$ significantly decreases when $\mathrm{f_c}$ is reduced below 0.1, which means that there is almost no noise in $\mathrm{LFS}_{noi}$.

\subsection{Image Multi-Frequency Decomposer}
\label{sec: Decomposer}

The ZSCFC theory leverages the consistency of image textures across multiple non-overlapping frequency bands to eliminate noise. Therefore, we first design an IMFD to iteratively decompose the noisy image in the frequency domain to obtain multiple frequency bands subimages, as illustrated in Fig. \ref{fig: frequency decomposer}. In ZSCFC, the proposed IMFD uses three experimentally determined cutoff frequencies $\mathrm{f_{c1}}$, $\mathrm{f_{c2}}$ and $\mathrm{f_{c3}}$ ($\mathrm{f_{c1}}\textless \mathrm{f_{c2}}\textless \mathrm{f_{c3}}\!=\!0.1$) to decompose a noisy image into four frequency bands, denoted as $\mathrm{LFS_1}$, $\mathrm{HFS_1}$, $\mathrm{HFS_2}$, and $\mathrm{HFS_3}$, as depicted in Fig. \ref{fig: 31low_freq_noise} (right) (values of $\mathrm{f_{c1}}$, $\mathrm{f_{c2}}$, $\mathrm{f_{c3}}$ are given in Section \ref{sec: implementation}). $\mathrm{LFS_2}$ and $\mathrm{LFS_3}$ are intermediate outputs obtained during the frequency decomposition process.

As shown in Fig. \ref{fig: frequency decomposer}, $\mathrm{LFS_1}$ is the lowest-frequency subimage, it is smooth and contains little noise. For the HFSs, due to the optimal $\mathrm{f_{c3}}$ (around 0.1) we have chosen, the highest-frequency $\mathrm{HFS_3}$ contains nearly all noise. In contrast, $\mathrm{HFS_1}$ and $\mathrm{HFS_2}$ have similar textures, with little noise presence.
\subsection{Cross-Frequency Consistency Loss}
\label{sec: 3.3}
For an $M\times N$ noisy image $img_{noi}=img+n$, where $n$ is zero-mean sensor noise of variance $\sigma^{2}_{n}$ and spatial correlation length $L_{c}$, and $img$ is the clean image. Denote by $\widehat{z} (\omega )$ the 2-D Fourier transform of any signal $z$, and by $\chi_{B} (\omega )$ the indicator of a radial band $B$. Fetching $n$ with the disjoint bands $B_{1}$ and $B_{2}$, and produces the noise components $X=\mathcal{F}^{-1}\!\bigl[\chi_{B_1}{n}\bigr]$ and $Y=\mathcal{F}^{-1}\!\bigl[\chi_{B_2}{n}\bigr]$. Because each correlation disc of area $\pi L^{2}_{c}$ overlaps only a fraction $\pi L^{2}_{c}/(MN)$ of the image, the cross-band covariance satisfies:

\begin{equation}
    Cov(X,Y)=\rho_{noise} \sigma^{2}_{n} ,\  \rho_{noise} \approx \frac{\pi L^{2}_{c}}{MN} \ll1
\end{equation}

So $X$ and $Y$ are virtually independent (e.g. $\rho_{noise}<10^{-3}$ for the images $L_c=3px$ and $256\times 256$). In contrast, Natural images have a Hölder-continuous spectrum $|\widehat{img} (\omega_{i} )-\widehat{img} (\omega_{j} )|\leq C_{img}||\omega_{i} -\omega_{j} ||^{\alpha }(0<\alpha \leq 1)$ with constant $C_{img}\sim 1$ and exponent $0<\alpha \leq 1$. Choosing co-radial frequencies $\omega_{1} \in B_{1}$ and $\omega_{2} \in B_{2}$, and Parseval’s theorem give almost-deterministic coupling between the band-limited textures $img_{1}=\mathcal{F}^{-1} \left[ \chi_{B_{1}} \widehat{img} \right]$  and $img_{2}=\mathcal{F}^{-1} \left[ \chi_{B_{2}} \widehat{img} \right]$:

\begin{equation}
    \rho_{tex} =\frac{Cov(img_{1},img_{2})}{\sigma_{img_{1}} \sigma_{img_{2}} } \geq 1-\frac{(C_{img}f^{\alpha }_{c,\max }\rho^{\alpha /2}_{12} )^{2}}{2\sigma^{2}_{img_{1}} } =1-\epsilon \approx 1
\end{equation}

where $f_{c,max}=0.1$ is the max cut-off frequency and $\rho_{12}=0.02$ is the relative gap between the two frequency rings, so $\epsilon <10^{-3}$.
Finally, a decisive gap can be shown by the ratio of texture-to-noise correlations:

\begin{equation}
\delta_{gap} =\frac{\rho_{tex} }{\rho_{noise} } \approx \frac{MN}{\pi L^{2}_{c}}(1-\epsilon) \gg 1    
\end{equation}

This gap means that a Cross-Frequency Consistency (CFC) loss can be designed and satisfy its denoising objective by eliminating the mutually uncorrelated noise while leaving the highly correlated texture.


\begin{figure*}[t]
  \centering
  \includegraphics[width=0.8\textwidth]{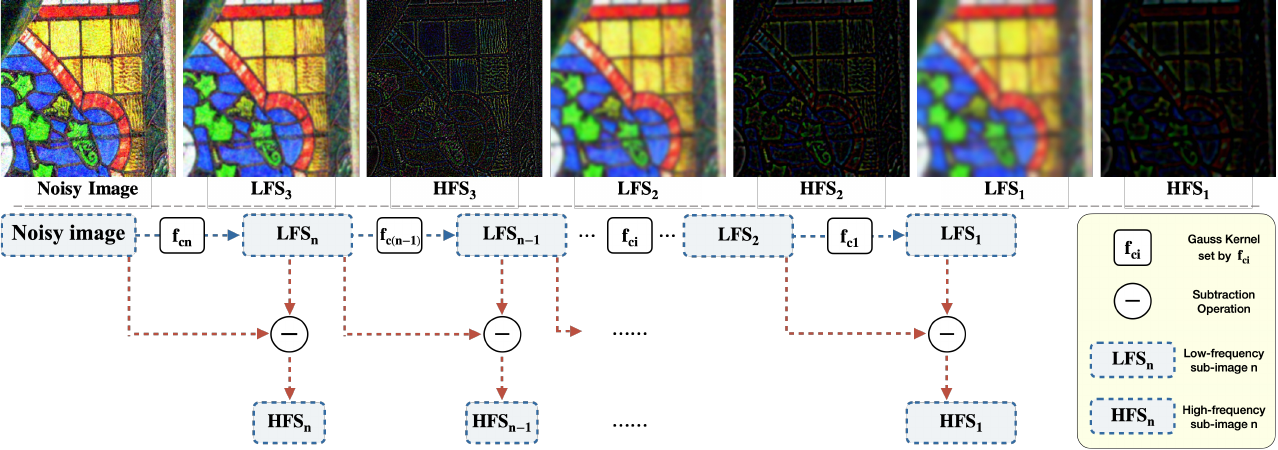}
  \caption{\textbf{Above:} An illustration of the output generated by the IMFD with $\mathrm{f_{c1}}$, $\mathrm{f_{c2}}$, and $\mathrm{f_{c3}}$. The $\mathrm{LFS_1}$, $\mathrm{HFS_1}$, $\mathrm{HFS_2}$, and $\mathrm{HFS_3}$ can be fused into the original noisy image. \textbf{Below:} Overview of our proposed IMFD framework.}
  \label{fig: frequency decomposer}
\end{figure*}


\noindent\textbf{Consistency Loss 1.}
Because $\mathrm{HFS_1}$ and $\mathrm{HFS_2}$ contain similar high-frequency textures, with minimal noise, we propose the first consistency loss function, $\mathcal{L}_{cons1}$, to guide the network in learning the distribution characteristics of high-frequency content. We aim to make the network extract as many high-frequency textures as possible from $\mathrm{HFS_1}$ and $\mathrm{HFS_2}$, so that when these extracted features undergo the decomposition process detailed in Fig. \ref{fig: frequency decomposer}, the result should be a composite image rich in textures. This implies that the fused image, after the extraction of textures, should satisfy the equations:

\begin{equation}
    \mathrm{LFS_2=LFS_1 +HFS_1, LFS_3=LFS_2 +HFS_2}
\end{equation}

\noindent Thus, we define $\mathcal{L}_{cons1}$ using the $L_1$-norm \cite{L1_loss} as follows:

\begin{equation}
\mathcal{L}_{cons1} = \mathrm{||LFS_1+g(HFS_1), LFS_3-g(HFS_2)||_1}
\end{equation}


\noindent\textbf{Consistency Loss 2.}
The second consistency loss function, $\mathcal{L}_{cons2}$ , employs two cutoff frequencies, a small $\mathrm{f_{ref1}}$ and a large $\mathrm{f_{ref2}}$, to produce a subimage of mid-frequency from the noisy image, $\mathrm{subimg_{ref}}$, as a texture reference. With a low $\mathrm{f_{ref1}}$ and a high $\mathrm{f_{ref2}}$, $\mathrm{subimg_{ref}}$ can include a substantial amount of textures (value of $\mathrm{f_{ref1}}$ and $\mathrm{f_{ref2}}$ are given in Section \ref{sec: implementation}). The generation of $\mathrm{subimg_{ref}}$ is shown in Fig. \ref{fig: mainArch}. We aim for the textures extracted by the network from $\mathrm{HFS_1}$ to $\mathrm{HFS_3}$, to align closely with $\mathrm{subimg_{ref}}$, thereby facilitating more comprehensive extraction of textures. To achieve this, each of the network texture extraction results $\mathrm{g(HFS_1)}$, $\mathrm{g(HFS_2)}$ , $\mathrm{g(HFS_3)}$ are compared to $\mathrm{subimg_{ref}}$ using the $L_2$-norm \cite{L1_loss}:


\begin{equation}
\mathcal{L}_{\text{cons2}}=||\mathrm{subimg_{ref}, g(HFS_3)}||_2 +\; ||\mathrm{subimg_{ref}, g(HFS_2)}||_2+\;||\mathrm{subimg_{ref}, g(HFS_1)||_2}
\end{equation}

\noindent\textbf{Regularization Loss.}
Maximizing texture extraction through $\mathcal{L}_{cons2}$ may inadvertently lead the network to extract noise. To mitigate this, a Total Variation (TV) regularization is used as regularization to help the network better distinguish between genuine textures and noise. Mathematically, for a given image $I\in \mathbb{R}^{H\times W}$, where $\mathrm{H}$ and $\mathrm{W}$ represent the height and width of the noise image, respectively, the TV regularization loss $\mathcal{L}_{reg}$ is expressed as:

\begin{equation}
\mathrm{
\bigtriangleup_{x} I\left( i,j\right)  =I(i,j)-I(i+1,j)
}
\end{equation}
\vspace{-10pt}
\begin{equation}
\mathrm{
\bigtriangleup_{y} I\left( i,j\right)  =I(i,j)-I(i,j+1)
}
\end{equation}
\vspace{-10pt}
\begin{equation}
\mathcal{L}_{reg} \mathrm{\left( I\right)  =\frac{1}{H\cdot W} \left( \sum^{H-1}_{i=1} \sum^{W}_{j=1} |\bigtriangleup_{x} I  |+\sum^{H}_{i=1} \sum^{W-1}_{j=1} |\bigtriangleup_{y} I  |\right)}
\end{equation}

Here, $\mathrm{\bigtriangleup_{x} I}$ and $\mathrm{\bigtriangleup_{y} I}$ measure the absolute differences between adjacent pixels along the horizontal and vertical directions respectively. 

\noindent\textbf{Total Loss.}
The total loss $\mathcal{L}_{total}$ is calculated by ($\omega_1$, $\omega_2$, $\omega_3$ are weight constants):
\begin{equation}
\mathcal{L}_{total} = \omega_1\mathcal{L}_{cons1} + \omega_2\mathcal{L}_{cons2} + \omega_3\mathcal{L}_{reg},
\end{equation}


\subsection{Ultralight Network}
\label{sec: Network}

Previous deep learning-based methods \cite{net1_uidfdk, net2_Jang_2023_ICCV, net3_mmbsn} often employed heavy networks to enhance the ability of feature learning, however, these methods can lead to overfitting and performance degradation when applied to single image denoising. Therefore, we designed an ultralight network with only approximately 1.5k parameters. 



\section{Experiments}

\begin{figure*}[t]
  \centering
  \includegraphics[width=0.8\textwidth]{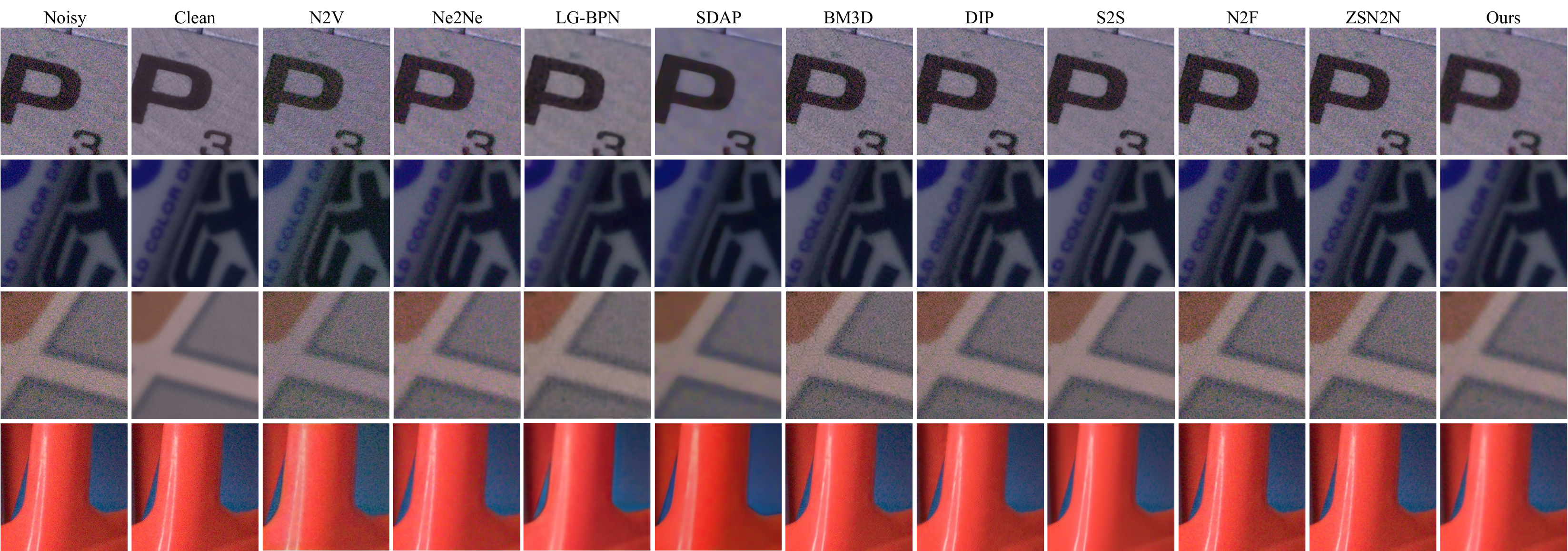}
  \caption{Visual quality comparison on SIDD Medium and Validation datasets.}
  \label{fig: exp}
\end{figure*}

\subsection{Implementation Details}
\label{sec: implementation}

\textbf{Training Details.} 
We implement our method with Python 3.8, PyTorch 1.13.1 on NVIDIA GeForce RTX 4090 GPUs. We employ two metrics to assess the denoising performance of the methods: the peak signal-to-noise ratio (PSNR) and the structural similarity index (SSIM). Higher PSNR and SSIM values indicate superior fidelity.

\noindent\textbf{Parameter Settings.} 
The $\mathrm{f_c}$ for IMFD are $\mathrm{f_{c3}=0.1}$, $\mathrm{f_{c2}=0.07}$, $\mathrm{f_{c1}=0.05}$, $\mathrm{f_{ref2}=0.12}$ and $\mathrm{f_{ref1}=0.03}$. The weights for $\mathcal{L}_{total}$ are $\omega_1=0.5$, $\omega_2=2$, $\omega_3=0.5$. The ablation study of parameter setting is given in the supplementary material.

\noindent\textbf{Datasets.} We conduct extensive experiments on six real-world datasets: RENOIR \cite{data1_renoir}, PolyU \cite{data2_polyu}, SIDD Medium \cite{data3_sidd}, SIDD Validation \cite{data3_sidd}, SIDD Benchmark \cite{data3_sidd}, and SenseNoise-500 \cite{data4_sense500} datasets, which have 120, 100, 320, 1280, 1280, 500 images respectively. Following the experimental settings of \cite{z4_ZSN2N}, we center-crop images from the RENOIR, PolyU, SIDD Medium, and SenseNoise-500 datasets into patches of size 256 $\times$ 256, ensuring consistency with the SIDD Validation and SIDD Benchmark datasets, which natively contain 256 $\times$ 256-sized images. These real-world datasets primarily lack images with high noise levels, thereby limiting the evaluation of our method across varying noise intensities. To address this and to demonstrate our method’s robustness under stronger noise conditions, we augment the Kodak24\footnote{http://r0k.us/graphics/kodak/} and McMaster18 \cite{data5_mc} datasets with synthetic pink noise, using $\mathrm{std}$ of 30 and 40 to simulate more substantial real-world noise. Pink noise is chosen due to its spectral distribution similarity to real-world noise, enhancing the realism of our simulated noise conditions. The noise level for all datasets is calculated by numpy, by $\mathrm{std(image_{noi}-image_{gt}})$. The SIDD Benchmark dataset does not provide ground-truth images, the noise level for this dataset is not reported in the table \ref{table: real-world}.

\noindent\textbf{Compared Methods.} We compare our ZSCFC with three zero-shot methods (DIP\cite{z1_DIP}, N2F\cite{z3_N2F}, ZSN2N\cite{z4_ZSN2N}), one traditional method (BM3D\cite{tr1_bm3d}), four self-supervised methods (N2V\cite{self1_n2v}, Ne2Ne\cite{self2_ne2ne}, SDAP\cite{self4_sdap}, LGBPN\cite{self3_lgbpn}). Due to the zero-shot method S2S \cite{z2_S2S} denoising requires over 40 minutes to process a single image, we consider its practical value to be limited. Therefore, we only tested S2S performance on a few randomly selected images in Section \ref{sec:4.4} for comparison.

\subsection{Real-World Experiments}

\begin{table*}[t]
\centering
\small
\renewcommand{\arraystretch}{0.7}
\caption{Quantitative comparison of ZSCFC and compared methods on six real-world image datasets in sRGB space. The highest PSNR(dB)/SSIM among the methods is marked in \textbf{bold}, while the second is \underline{underlined}.}
\resizebox{0.85\textwidth}{!}{%
\begin{tabular}{cc|cccc|ccccc}
\toprule
\multirow{2}{*}{Dataset} & \multirow{2}{*}{Metric} & \multicolumn{4}{c|}{Self-supervised methods} & \multicolumn{5}{c}{Zero-shot methods} \\
\cmidrule{3-11}
& & N2V & Ne2Ne & LG-BPN & SDAP & BM3D & DIP & N2F & ZSN2N & \textbf{Ours} \\
\midrule
RENOIR & PSNR & 27.61 & 28.68 & \underline{31.11} & 30.03 & 28.88 & {29.12} & 28.74 & 28.64 & \textbf{33.31} \\
$\mathrm{std}$ = 12.12 & SSIM & 0.617 & 0.611 & 0.776 & \underline{0.784} & 0.650 & {0.674} & 0.608 & 0.600 & \textbf{0.798} \\
\midrule
PolyU & PSNR & 30.18 & 36.22 & \underline{36.61} & 28.89 & 35.92 & {36.27} & 36.19 & 36.13 & \textbf{37.28} \\
$\mathrm{std}$ = 4.107 & SSIM & 0.891 & 0.920 & 0.928 & 0.913 & 0.909 & \underline{0.942} & 0.916 & 0.911 & \textbf{0.949} \\
\midrule
SIDD Medium & PSNR & 27.60 & 30.17 & 31.23 & 28.60 & 30.50 & \underline{31.78} & 29.96 & 29.92 & \textbf{35.15} \\
$\mathrm{std}$ = 11.73 & SSIM & 0.639 & 0.676 & 0.778 & \underline{0.865} & 0.753 & 0.747 & 0.659 & 0.645 & \textbf{0.874} \\
\midrule
SIDD validation & PSNR & 25.10 & 26.33 & \underline{30.54} & 28.93 & 28.77 & 26.63 & 25.59 & 25.61 & \textbf{32.59} \\
$\mathrm{std}$ = 18.90 & SSIM & 0.406 & 0.470 & 0.765 & \textbf{0.809} & 0.709 & 0.509 & 0.435 & 0.423 & \underline{0.773} \\
\midrule
SIDD benchmark & PSNR & 31.00 & 31.23 & \underline{32.54} & 30.93 & 29.63 & 31.21 & 30.62 & 30.19 & \textbf{34.33} \\
-- & SSIM & 0.751 & \underline{0.794} & {0.793} & 0.785 & 0.740 & 0.528 & \textbf{0.878} & 0.429 & 0.773 \\
\midrule
SenseNoise-500 & PSNR & - & - & - & - & \underline{26.73} & {26.50} & 26.00 & 25.91 & \textbf{27.95} \\
$\mathrm{std}$ = 17.29 & SSIM & - & - & - & - & 0.523 & \underline{0.573} & 0.559 & 0.546 & \underline{0.690} \\
\bottomrule
\end{tabular}%
}
\label{table: real-world}
\end{table*}

\textbf{Details of Real-World Experiments.} 
The zero-shot methods were directly applied to each image for denoising purposes. The traditional denoiser, BM3D, requires an estimated noise level ($\sigma$) as a parameter; thus, we employed the optimal noise estimation method \cite{tr2_bm3d_op} for BM3D. 
For self-supervised methods, these methods are trained on SenseNoise-500 dataset and applied to denoise other datasets in line with their experimental settings.

\noindent\textbf{Results of Real-World Experiments.} 
Table \ref{table: real-world} shows the quantitative comparison of six real-world datasets. Our ZSCFC method has achieved the best denoising performance in PSNR across six real-world datasets. Since these datasets contain noise samples with std ranging from 4 to 20, the adaptability and robustness of our method were demonstrated in both low and high real-world noise scenarios. In comparison, ZSN2N performed well only with lower noise levels\footnote{In the ZSN2N paper, their real-world denoising experiments mention that ``we randomly choose 20 images from both datasets to test on''. We believe that this may lead to biased results, so we evaluated all images in the dataset for a fairer comparison, explaining the discrepancy in PSNR between our results and those in Table \ref{table: real-world}.}. This may be due to the overlap of noise across the downsampled images when noise levels are high. With only a single downsampling, ZSN2N struggled to leverage the independence of noise, which limited the network’s ability to extract noise features effectively. Similarly, N2F, derived from N2N theory, faced the same limitations, highlighting the challenges of achieving both effectiveness and efficiency with N2N theory in zero-shot denoising. DIP relies on its early stopping mechanism, stopping too early can lead to incomplete denoising, while stopping too late can result in a loss of image textures. Although BM3D was provided with optimal parameters, its performance still lagged significantly behind our method. Lastly, dataset-dependent self-supervised methods all performed poorly when trained and denoised with different datasets, indicating that these methods are not directly applicable to single-image denoising tasks. Furthermore, Fig. \ref{fig: exp} shows the superiority of our method over competing methods. Our ZSCFC recovers more textures and has a higher degree of noise removal.

\subsection{Synthetic Experiments}

\begin{table*}[t]
\centering
\small 
\renewcommand{\arraystretch}{0.6} 
\caption{Quantitative comparison of ZSCFC and compared methods for synthetic pink noise.}
\resizebox{0.8\textwidth}{!}{%
\begin{tabular}{cc|cccc|ccccc}
\toprule
\multirow{2}{*}{Dataset} & \multirow{2}{*}{Metric} & \multicolumn{4}{c|}{Self-supervised methods} & \multicolumn{5}{c}{Zero-shot methods} \\
\cmidrule{3-11}
& & N2V & NB2NB & LG-BPN & SDAP & BM3D & DIP & N2F & ZSN2N & \textbf{Ours} \\
\midrule
McMaster18 & PSNR & 14.77 & 16.59 & 18.65 & 15.01 & 19.26 & 20.33 & \underline{21.28} & 20.97 & \textbf{21.55} \\
$\mathrm{std}$ = 28.95 & SSIM & 0.409 & 0.475 & 0.374 & 0.353 & \underline{0.554} & 0.470 & 0.548 & 0.519 & \textbf{0.595} \\
\midrule
McMaster18 & PSNR & 14.28 & 16.13 & 16.53 & 14.62 & 17.70 & 17.83 & \underline{19.19} & 18.99 & \textbf{19.56} \\
$\mathrm{std}$ = 35.96 & SSIM & 0.386 & 0.461 & 0.298 & 0.360 & \underline{0.478} & 0.372 & 0.476 & 0.450 & \textbf{0.521} \\
\midrule
Kodak24 & PSNR & 16.58 & 18.29 & 18.46 & 15.73 & 20.81 & 20.27 & \underline{21.21} & 21.17 & \textbf{21.56} \\
$\mathrm{std}$ = 29.90 & SSIM & 0.479 & 0.529 & 0.372 & 0.384 & \textbf{0.601} & 0.458 & 0.541 & 0.529 & \underline{0.593} \\
\midrule
Kodak24 & PSNR & 15.35 & 16.82 & 15.57 & 14.62 & 17.63 & 16.78 & \underline{18.11} & 18.07 & \textbf{18.51} \\
$\mathrm{std}$ = 41.91 & SSIM & 0.450 & \textbf{0.500} & 0.281 & 0.367 & \underline{0.482} & 0.335 & 0.435 & 0.428 & 0.479 \\
\bottomrule
\end{tabular}%
}
\label{table: Pink Result}
\end{table*}

\textbf{Details of Synthetic Experiments.} 
Pink noise, characterized by its 1/$\mathrm{f}$ spectral decay with frequency $\mathrm{f}$, is a type of nonlinearly decaying noise with an uneven spectral distribution, posing significant challenges for removal. This noise type can somewhat replicate the randomness in real noise spectrum distributions. We generated two different intensities of pink noise with $\mathrm{std}$ set at 30 and 40. 

\noindent\textbf{Results of Synthetic Experiments.}
Table \ref{table: Pink Result} presents the quantitative comparison for synthetic pink noise. Our method achieves the highest performance, significantly surpassing both zero-shot and self-supervised approaches. These results further validate the superior performance of our method across different noise intensities.

\subsection{Computational Efficiency Experiments}
\label{sec:4.4}

We randomly selected five images from the SIDD Medium dataset and applied the zero-shot methods to each single image. The average inference time required for denoising a single image was measured. As shown in Fig. \ref{fig: page1}(b) and Table \ref{tab: ablation}(a), our ZSCFC method has an order-of-magnitude advantage in terms of parameter count and GFLOPs, with inference time reduced to half that of ZSN2N (the second best). 

\begin{table*}[t]
  \centering
  \caption{\textbf{(a)} Quantitative comparison of Computational Efficiency. First Row: Non-learning based method. Second to sixth Rows: Learning based method. \textbf{(b)} Ablation study on depth of the network. \textbf{(c)} Ablation study on loss function.}
  \includegraphics[width=0.8\textwidth]{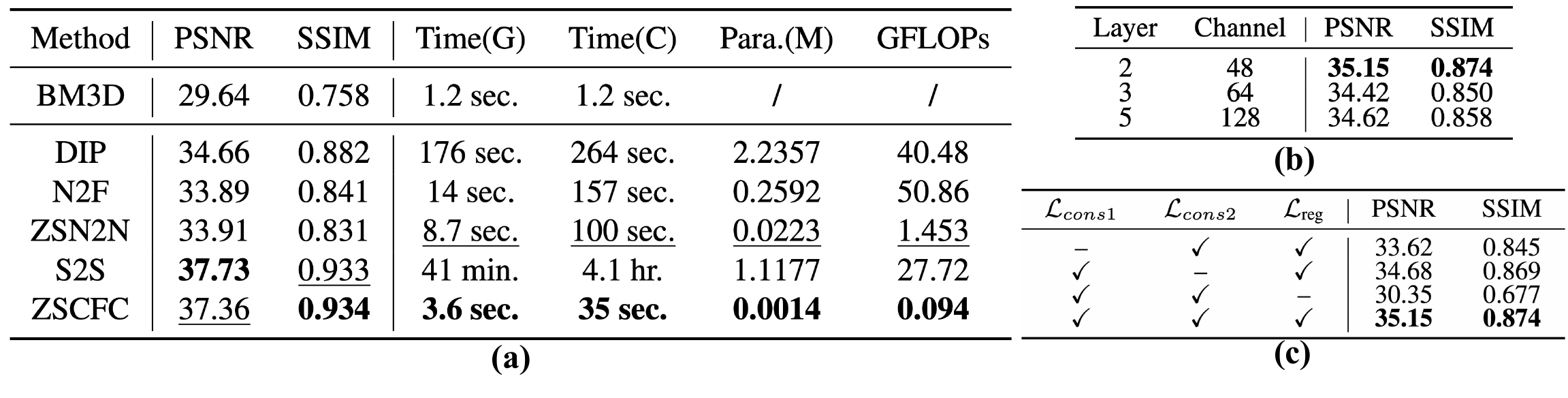}
  \setlength{\abovecaptionskip}{-0.05cm}
  \label{tab: ablation}
\end{table*}

\subsection{Ablation Study}

We conducted ablation studies to analyze the influence of the depth of the network and the loss function on the SIDD Medium dataset. In addition, the ablation study of the hyperparameters is provided in the Supplementary Material due to space limitations.

\noindent\textbf{Depth of Network.} 
As shown in Table \ref{tab: ablation}(b), we experimented with increasing the depth of the network to three and five layers. Both the three-layer and five-layer networks exhibited overfitting, resulting in reduced denoising performance.


\noindent\textbf{Loss Function.}
We evaluated the necessity of the two consistency loss functions and the smoothness regularization term. Table \ref{tab: ablation}(c) represents the cases where $\mathcal{L}_{cons1}$, $\mathcal{L}_{cons2}$, $\mathcal{L}_{reg}$ are omitted, respectively. The absence of each loss function resulted in a decrease in denoising performance, confirming the positive contribution of each loss term to guiding the network to learn high-frequency image information effectively.


\section{Conclusion}

In this paper, we propose ZSCFC, a zero-shot image denoising method designed for real-world scenarios. By utilizing cross-frequency consistency, our method effectively guides an ultralight network to extract textures from an image and complete denoising tasks. The proposed network has only 1.5k parameters and requires just 3 seconds of GPU processing time per image. Despite its compact size, ZSCFC outperforms larger networks with millions of parameters, demonstrating its suitability for deployment on edge devices with limited computational resources.

\bibliographystyle{unsrt}  
\bibliography{templateArxiv}

\end{document}